# Design of Efficient Deep Learning models for Determining Road Surface Condition from Roadside Camera Images and Weather Data

Juan Carrillo[1], Mark Crowley[1], Guangyuan Pan[2], Liping Fu[2]
1. Department of Electrical and Computer Engineering, University of Waterloo
2. Department of Civil and Environmental Engineering, University of Waterloo

## Abstract

Road maintenance during the Winter season is a safety critical and resource demanding operation. One of its key activities is determining road surface condition (RSC) in order to prioritize roads and allocate cleaning efforts such as plowing or salting. Two conventional approaches for determining RSC are: visual examination of roadside camera images by trained personnel and patrolling the roads to perform on-site inspections. However, with more than 500 cameras collecting images across Ontario, visual examination becomes a resource-intensive activity, difficult to scale especially during periods of snowstorms. This paper presents the results of a study focused on improving the efficiency of road maintenance operations. We use multiple Deep Learning models to automatically determine RSC from roadside camera images and weather variables, extending previous research where similar methods have been used to deal with the problem. The dataset we use was collected during the 2017-2018 Winter season from 40 stations connected to the Ontario Road Weather Information System (RWIS), it includes 14.000 labeled images and 70.000 weather measurements. We train and evaluate the performance of seven state-of-the-art models from the Computer Vision literature, including the recent DenseNet, NASNet, and MobileNet. Moreover, by following systematic ablation experiments we adapt previously published Deep Learning models and reduce their number of parameters to about ~1.3% compared to their original parameter count, and by integrating observations from weather variables the models are able to better ascertain RSC under poor visibility conditions.

## Introduction

Road maintenance during the Winter is a safety critical operation that requires a significant amount of resources. In countries located in Northern latitudes, such as Canada, The United States, and Finland, Winter road maintenance is a priority for Government Offices at multiple levels. While these countries have made substantial efforts to ensure that the roads are suitable for the transportation of passengers and goods during the Winter, achieving a tradeoff between safety and resources expended is a challenge they face every year.

Road safety

Driving when there is snow or ice on the roads is not only more difficult but also more dangerous. The harsh weather conditions during the Winter are recognized as key factors associated with a higher probability of collisions [1] [2] due to circumstances such as poor road surface friction [3]. Moreover, the risk of fatality also increases in accidents occurring in periods of snow falls or even after snow falls if the road surface has not been cleaned timely and thoroughly [4]. Transportation offices are aware of these numbers and plan accordingly, with specific measures to prioritize road monitoring and cleaning as part of their short-term plans and long-term road safety strategies.

Being one of the largest countries in the Northern hemisphere, Canada deals with Winter road maintenance operations over thousands of kilometers of urban and rural roads every year. Even though it is a country with experience on how to keep its roads operative in the Winter, there is still room for improvement in terms of road safety. For instance, in the year 2013, more highway fatalities due to snow or ice conditions were reported in the province of Ontario in Canada compared to the previous year [5]. On the positive side, also in Ontario the benefits of a timely winter road maintenance have been quantified [6], highlighting the importance of data-driven methods for improving operations.

Road maintenance costs and operations

The cost of Winter road maintenance is significantly higher than that spent in other seasons of the year. For instance, according to the Ministry of Transportation in Ontario (MTO), about 50% of the total budget disbursed for highway maintenance corresponds to snow and ice cleaning during the Winter [7]. In Montreal, the budget for snow and ice removal is approximately 150 million CAD annually [8], with similar figures in Toronto (90 million CAD) [9] and Ottawa (68 million CAD) [10].

In Canada, every province has autonomy in how they manage Winter road maintenance operations. Depending on the strategy established by the Transportation offices at the provincial or municipal level, the maintenance is done directly by the Transportation offices or by private contractors. Moreover, the standards for maintaining road surface condition during the Winter also differ across provinces according to their particular needs and environmental conditions [8]. However, in a recent report published by the Office of the Auditor General of Ontario, road safety concerns were raised due to the outsourcing of snow and ice removal operations to private contractors, including observations about the lack of a proper system to oversee snow removal operations and tools to monitor road surface cover during snow falls [5].

Multiple activities need to be scheduled and coordinated by the Transportation offices in order to keep roads with the minimal possible amount of snow and ice so the drivers can use the road network safely. For this purpose, the transportation offices commonly use Road Surface Condition (RSC) as a measure to identify the current state of the road regarding snow or ice coverage, serving also as a communication mechanism between stakeholders across the whole operation. More specifically, RSC introduces several categories to quantify the current amount of snow cover over a road. In Ontario, the Ministry of Transportation (MTO) uses three main categories to identify whether the road is bare, partly covered, or fully covered [11]. Examples of images for each category are shown in Figure 1.

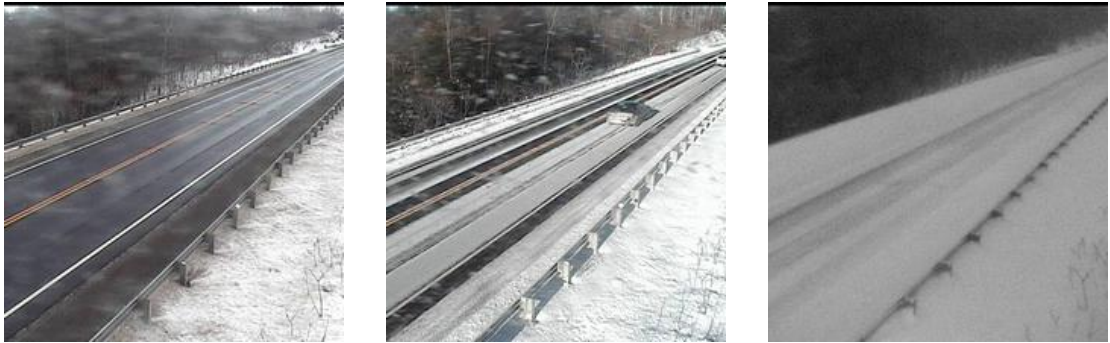
Figure 1. Example images from a roadside camera near Otter lake in Ontario.
Left: Bare pavement. Center: Partial snow cover. Right: Full snow cover.

In terms of logistics, the overall Winter road maintenance operation commonly involves two major stages. First, a road monitoring stage continuously checks road surface coverage when snowfall is highly likely to occur, then, once the snowfall has stopped a cleaning stage begins and the necessary resources, such as plowing and salting, are assigned to promptly remove snow or ice accordingly. The road monitoring stage commonly involves patrolling by trained inspectors as well as observation through roadside cameras and weather stations. Even though patrolling is the most accurate method to identify RSC, it is also costly, and its coverage is limited to only those roads traveled by the inspectors.

Road Weather Information Systems (RWIS)

The Road Weather Information System (RWIS) was introduced as a network of advanced stations to continuously monitor roads at certain locations, improving the coverage and complementing the conventional road patrolling [7]. In general, RWIS stations include roadside cameras along with weather sensors to measure variables such as air temperature, wind speed, pressure, and humidity. Depending on the vendor and the configuration, the stations can also have embedded pavement sensors to determine road surface temperature.

Previous research related to RWIS stations has focused primarily on quantifying their overall benefits for road Winter maintenance, both in terms of safety and cost savings [12] [13], as well as best practices for determining their optimal locations over an area of interest [14] [15]. The evidence on previous studies confirms the relevance of RWIS stations for Winter maintenance operations, with networks of stations installed across North America and Northern Europe. However, with dozens or even hundreds of stations collecting data 24 hours per day, it is difficult for the operators of those sensor networks to process all that data simultaneously and perform decision making in almost real time during snowfalls.

Automated processing of RWIS stations data is potentially beneficial for multiple stakeholders involved in the Winter road maintenance operations. On one hand, private companies hired by Transportation offices monitor the roads using the RWIS network and send salt trucks and snowplows to maintain road surface standards when needed. On the other hand, Transportation offices also monitor the roads through the RWIS network to oversee the private contractor's work and publish road status information to the general public.

Deep Learning (DL) methods are a subset of Machine Learning (ML) methods resulting from a combination of advanced algorithms and math [16]. The remarkable achievements of DL methods across multiple applications are also attributed to the recent availability of powerful computing hardware, mainly Graphic Processing Units (GPUs) to train DL models. One of the fields where DL methods have achieved remarkable results is in Computer Vision, showing state-of-the-art accuracy in tasks such as image classification, object detection, and semantic segmentation. In most of the major Computer Vision research conferences, such as CVPR, ICCV, and IGARSS, DL methods have consistently shown better performance across applications like Medical Imaging, Autonomous Vehicles, Remote Sensing, Face Recognition, among other fields.

Recently, a handful of studies have evaluated the use of DL to automate the classification of images with the purpose of determining RSC during the Winter, with remarkable results over images from dash cameras [17][18][19][20]. However, fewer studies have tested DL methods for determining RSC from images collected by roadside cameras. One example is the work of Pan et. al. [21], in which they adapt DL models trained over the ImageNet dataset to the task of RSC classification using the fine-tuning technique, achieving an accuracy of more than 90%.

Research objectives and contributions

The objectives of this study are twofold. First, we aim to extend previous work regarding the use of DL methods for automated classification of RSC using images from roadside cameras with a particular focus on evaluating state-of-the-art image classification methods such as DenseNet [22], NASNet [23], and MobileNet [24]. Secondly, we want to determine experimentally if the use of weather variables as a complementary data source improves the classification accuracy.

This paper is divided into three major sections. The first section describes the area of study as well as the datasets we use. In the second section, we evaluate multiple DL models for classification of RSC and in the last part, we include weather data as an additional source of data and assess the improvement on classification accuracy. To the best of our knowledge, this is the first study that looks at the combined use of roadside camera images and weather data for the automated determination of RSC. Additional contributions of this study include Python source code to replicate all the experiments presented in the paper. See

**Area of study and data**

The area of study comprises the center and south part of the province of Ontario in Canada, which hosts approximately 38% of the population in the North American country. The road network in the province extends along more than 275.000 km including all categories of roads, such as highways and municipal roads [25]. However, most of the population and therefore the roads are in the southern area. Weather conditions can vary drastically across the province, due mainly to the large range of latitude it covers, from approximately 42° N to 56° N, and the weather effects caused by the Great Lakes in its south-west side. For this reason, about 140 RWIS stations have been installed by the Ministry of Transportation in Ontario to help monitor the road conditions and weather in carefully selected locations across the province.

For this study, we selected a sample of 40 RWIS stations shown in Figure 2, which collected data approximately every 15 minutes for the Winter of 2017-2018. The data includes 14.000 images taken by the roadside cameras as well as 70.000 observations from five weather variables measured by instruments installed on every station. Every image was labeled according to one of the three RSC categories listed by the Ministry of Transportation in Ontario, with approximately 45% of images corresponding to bare pavement, 40% to partial snow cover, and 15% to full snow cover. The recorded weather variables are Air Temp (°C), Relative Humidity (%), Pressure (kPa), Wind Speed (km/h), and Dew Point (°C).

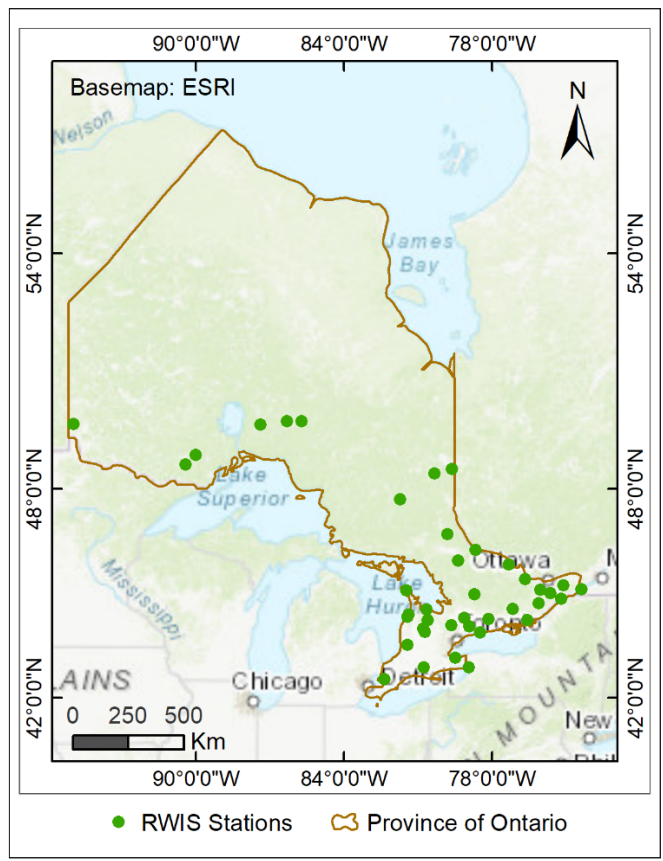

Figure 2. Location of the sample 40 RWIS stations in the province of Ontario.

**Classification of Road Surface Condition with Deep Learning models**

DL methods have achieved remarkable accuracy for tackling multiple Computer Vision tasks in both research and industry settings, with applications ranging from automated diagnosis in medical imaging to quick detection of cyclists and pedestrians in autonomous vehicles, just to name a few. One of the factors that have influenced the rise of DL is the current availability of large datasets, such as ImageNet [26] with millions of images labeled according to predefined classes, which are essential for training DL algorithms. However, having such large datasets is not common for industry-specific applications; therefore, researchers have developed a practice called fine-tuning in which DL models trained over large and generic image datasets are adapted for classification of images in other domains.

DL models are an evolution of a simpler type of model called the Multi-Layer Perceptron, which takes an input vector and uses consecutive groups of non-linear functions called layers to produce higher-level representations of the input data. In contrast, DL models use many layers, each one taking as input the output of the previous one. Most DL models fall into the category of supervised learning because their goal is to make the model create a mapping from input observations into the desired output, in our particular case we expect the DL models to receive images from roadside cameras as input and output the correspondent RSC category as defined by MTO. Figure 3 shows a simplified description of the architecture of DL models for image classification. An input image is represented as a three-dimensional matrix, with the height, width, and Red Green Blue (RGB) channels of the image as its initial dimensions. This input image moves through the layers in the model and in the end, the model outputs a vector of probabilities, with the highest probability corresponding to the most likely RSC category for the image.

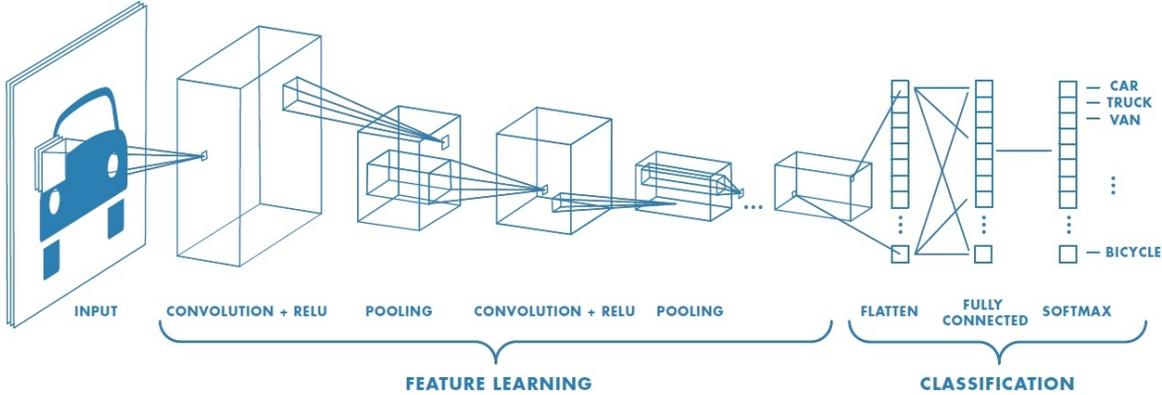

Figure 3. The generic architecture of a DL model for image classification. Image source: mathworks.com

The first part of DL models for image classification (feature learning) includes a series of convolutional and pooling layers that gradually convert the information contained on the input image into a compressed vector representation with smaller height and width but a larger number of channels than the original image. Convolutional layers are also known as filters that summarize an input matrix into one with smaller dimensions. In other words, these layers learn how to detect features such as shapes, contrast patterns, and color variations, and pass along a simplified representation of this features to the following layers.

The second part (classification) starts with a layer (flatten) that converts the three-dimensional vector from the first section into a one dimensional vector that is then feed into a few fully connected layers that reduce even more the size of the vector representation, and finally a softmax layer that outputs probability values for each category in the classification task. Fully connected layers are usually implemented in the last part of the models to summarize the visual features detected by previous layers. Most DL models have many convolutional layers but only a few fully connected layers because the number of parameters in the latter ones is much higher, which in turn requires more computational resources to train the models.

The design of a DL model also includes the consideration of multiple hyperparameters, and their role is crucial for the successful implementation of the model. The designer can set hyperparameters to manage multiple aspects of the model and fine-tuning them commonly requires significant effort and domain knowledge of the application. By setting those parameters the designer can define the number of neurons per layer, the type of non-linear functions to use (activation functions), the number of times the dataset is passed through the model during training (epochs), how fast the model learns (learning rate), the Dropout regularization rate, among other characteristics of the model definition and training. In practice, researchers focus on fine-tuning the most relevant hyperparameters based on previous literature and their experimental findings.

DL models to compare

We select six DL models that have scored state-of-the-art accuracy over the ImageNet image classification benchmark and evaluate how they perform for the classification of RSC over images from roadside cameras. All six models have been previously trained using the ImageNet dataset, which includes images of everyday objects such as animals, artifacts, and plants, among other classes. The goal is to fine-tune or retrain only a small portion, generally the latest portion, of those models in order to adapt them into the RSC image classification task. The underlying hypothesis is that the initial part of those DL models trained on large datasets has learned to identify basic patterns that are useful for classifying objects across multiple domains.

The DL image classification models we choose are Inception-v3 [27], Inception-Resnet-v2 [28], Xception [29], DenseNet169 [22], MobileNetv2 [24], and NASNet [23]. For all these models we kept the configuration of layers, training, hyperparameters, and regularization techniques as published in the original papers. When running the fine-tuning process, we consider previous settings from the work by Pan et. al. [21] who also explored the use of DL models for RSC classification, as well as other literature for fine-tuning image classification models [30][31][32].

In addition, we design a rather simple model inspired by seminal works in DL such as AlexNet [33] and VGG [34]. This baseline model has approximately 3.7% of the number of layers and 4.2% of the number of parameters when compared with the average characteristics of the other six models. Table 1 presents a summary of the characteristics of all seven models considered. Layers used for data reduction (max-pooling), model regularization (dropout), and vector concatenation are also counted as part of the total number of layers.

Table 1. Main characteristics of the DL models selected for comparison.

| Model | Feature learning part | | Classification part | | Complete model | |
|---|---|---|---|---|---|---|
| | # Parameters | # Layers | # Parameters | # Layers | # Parameters | # Layers |
| Baseline | 392,608 | 10 | 603,411 | 7 | 996,019 | 17 |
| Inception-v3 | 21,802,784 | 311 | 6,292,755 | 7 | 28,095,539 | 318 |
| Inception-Resnet-v2 | 54,336,736 | 780 | 4,719,891 | 7 | 59,056,627 | 787 |
| Xception | 20,861,480 | 132 | 9,831,699 | 7 | 30,693,179 | 139 |
| DenseNet169 | 12,642,880 | 595 | 3,915,027 | 7 | 16,557,907 | 602 |
| MobileNetv2 | 2,257,984 | 155 | 62,739 | 7 | 2,320,723 | 162 |
| NASNetMobile | 4,269,716 | 769 | 51,987 | 7 | 4,321,703 | 776 |

Model training

All seven models are trained and validated using 80% of the total number of images (14.000). The remaining 20% is held as a test set and we only use it for reporting accuracy in the end. Within the 80% of training data, 20% is taken for validation and plotting of the accuracy and loss functions during the training stage. We use the Backpropagation algorithm and Stochastic Gradient Descent (SGD) as optimizer to minimize the classification error (loss). Learning rate is kept constant at 0.001. Other parameters for the SGD optimizer are set according to recommended values in the literature, such as momentum = 0.9 and Nesterov momentum enabled. The number of times (epochs) the training set is passed through each DL model to iteratively reduce the misclassification error is kept constant at 50 and the images are feed in groups of 32 (batch size) to reduce memory usage.

Since the models have been previously trained using the ImageNet dataset, we only need to fine-tune or retrain the last portion of them in order to adapt them to the RSC classification task, except for the baseline model that needs to be trained from scratch. We consider three fine-tuning scenarios in which different percentages of the models are retrained, starting with only the fully connected layers, then fine-tuning only the last 5% and 15% of the models, including some of the last convolutional layers. During the training, a validation set is kept apart to continuously evaluate how the model performs over unseen images.

Results

Figures 4, 5, and 6 summarize the training and validation accuracy scored by the models when fine-tuning the fully connected layers and the last 5% and 15% percent of the models, except for the baseline model in which all the layers are trained from scratch. To our surprise, the baseline model achieved a competitive accuracy while all six state-of-the-art models show signs of overfitting, which means they learned successfully how to classify images in the training set but are unable to correctly classify images not seen during the training stage.

Overfitting is a common issue in Machine Learning, and it happens in cases where the model is too complex for a task and as a result, it memorizes the characteristics of the training set. In other words, the model does not generalize well for data samples not observed during training, like the images in the validation or the test sets. Multiple regularization techniques are available to avoid or mitigate overfitting in DL models, such as reducing the number of layers, reducing the number of neurons, or using Dropout layers that randomly *disconnect* certain connections between neurons, among others.

To avoid overfitting in our study, we use three Dropout layers with a rate of 0.5 between the fully connected layers of all models. The rate defines the frequency at which the connections between neurons in those layers are randomly dropped, the higher the rate the stronger the regularization effect; however, using a rate that is too high could hinder the ability of the model to learn any meaningful patterns during training.

The number of Dropout layers as well as the rate we use are appropriate according to the practices commonly found in the literature for these type of DL models. It is worth to note that each of the DL models we used from the literature already includes their own strategies to mitigate overfitting; therefore, we do not apply additional measures. Instead, we introduce the baseline model with the goal of evaluating the performance of a much simpler architecture with a significantly smaller number of layers and parameters.

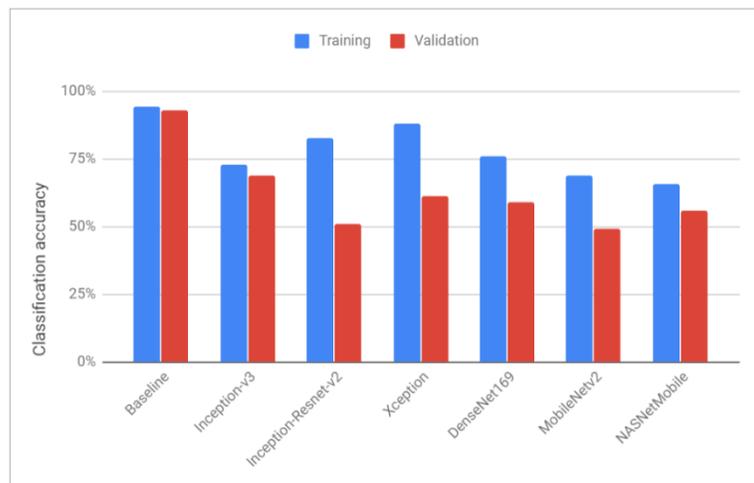

Figure 4. Finetuning only the fully connected layers.

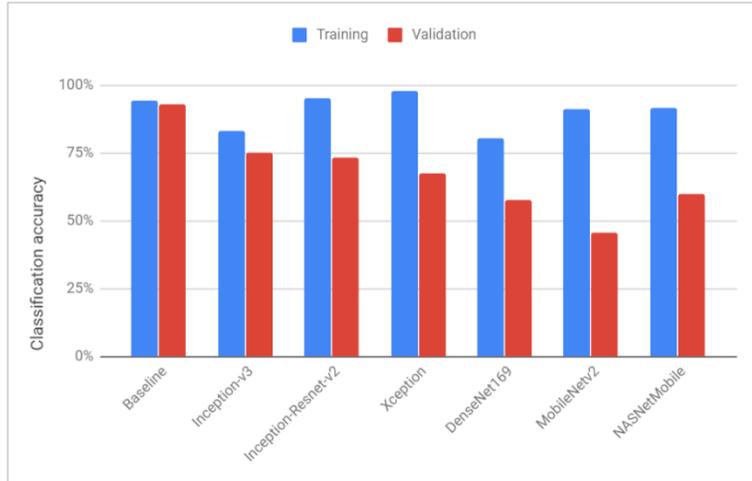

Figure 5. Finetuning the last 5% of layers.

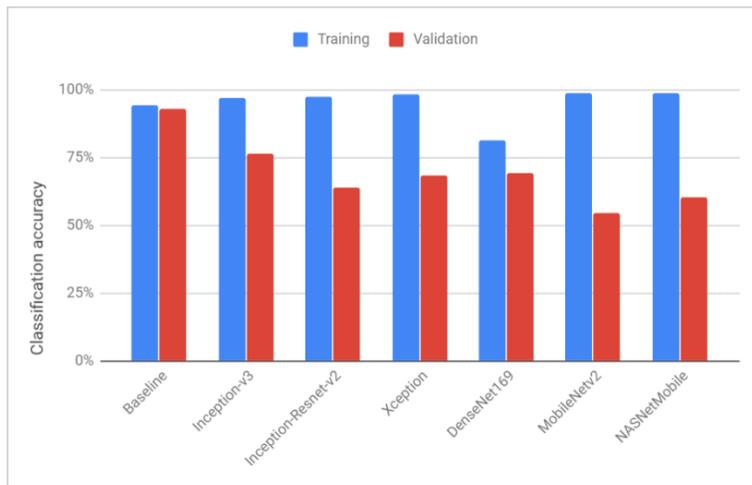

Figure 6. Finetuning the last 15% of layers.

Increasing the number of fine-tuned layers for the RSC classification task produces higher accuracy in the training set; however, it also affects considerably the accuracy on the validation set. In other words, the more layers we fine-tune the stronger the overfitting effect, with extreme cases such as the MobileNetv2 model, which scores 98.59% training accuracy but only 54.44% validation accuracy when the last 15% of the model is retrained. On the other hand, the baseline model scores more than 93% classification accuracy in both the training and validation sets. To better illustrate these findings, we show the variation of the loss and accuracy functions during the training of the MobileNetv2 model in Figure 7 and the baseline model in Figure 8.

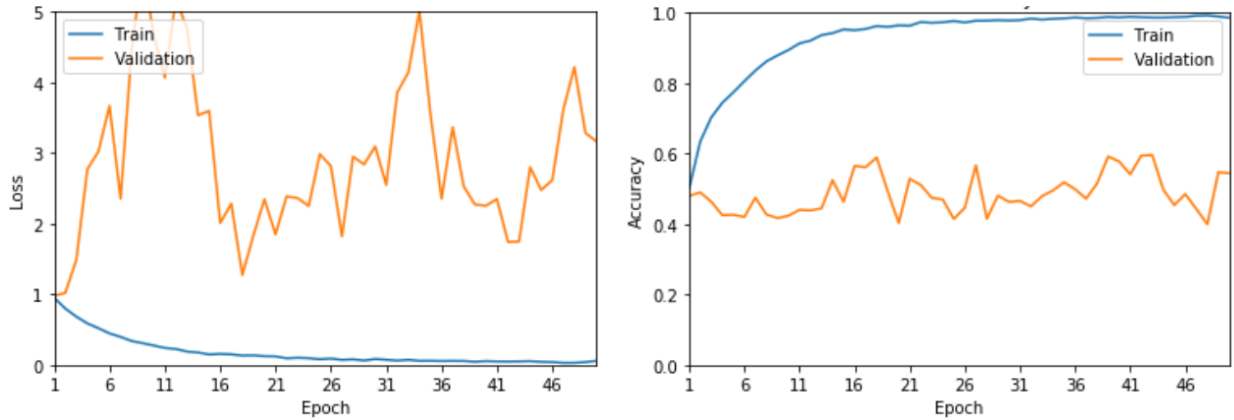
Figure 7. Fine-tuning the last 15% of the MobileNetv2 model. Left: Loss function. Right: Accuracy.

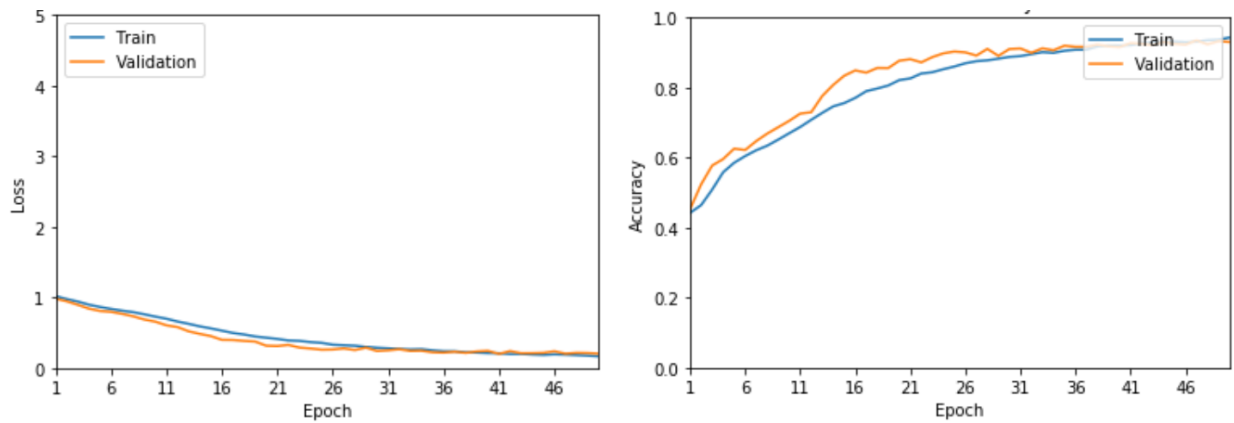
Figure 8. Training of the baseline model. Left: Loss function. Right: Accuracy.

On average, the difference between the final training and validation accuracy for the six state-of-the-art models is 18.2% when finetuning only the fully connected layers, 26.6% when finetuning the last 5% of layers, and 29.7% when finetuning the last 15% of layers. In contrast, the difference between the final training and validation accuracy for the baseline model is just 1.3%.

To summarize, the desired performance for any Machine Learning model is to have high accuracy in both the training and validation sets; however, in our experiments, this goal was only achieved by the baseline model. While multiple explanations can clarify these results, the most reasonable one is that the RSC image classification task might not require those complex DL models and a competitive accuracy can be scored using a simpler DL model like the baseline.

Ablation study

We also investigate how much we can simplify the baseline model without affecting the resulting classification accuracy. Experiments that remove some parts of a Machine Learning model and look at the effects it produces on the model performance are known as ablation studies and are recommended especially for DL models [35][36]. We explore two different ways to reduce the complexity of the baseline model while maintaining its classification accuracy. The first is focused on reducing the number of channels in the intermediate convolutional layers (feature learning part) and the second looks at reducing the number of neurons in the fully connected layers (classification part). Figure 9 shows the initial architecture of our baseline model.

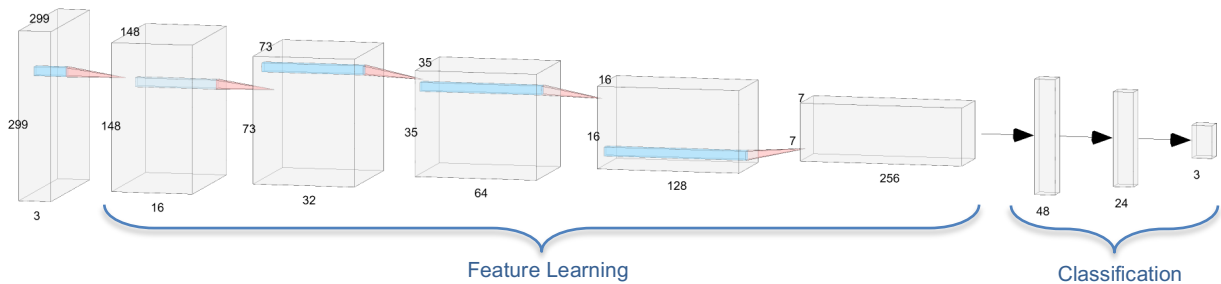

Figure 9. The architecture of the baseline model.

In this model, the number of channels doubles from one convolutional layer to the following, the same as in the architecture of VGG [34]. We call the rate at which the number of channels grows as Incremental Channels Factor (ICF) and explore the effect of using an ICF smaller than two. Figure 10 shows the effects of reducing the ICF in the baseline model where the area of the circles is representative of the total number of parameters in the model. Lowering the ICF reduces the validation accuracy; however, the decrement is not significant when the value is kept equal or above ~1.7; therefore, we move forward using this number. Using a smaller ICF also reduces the number of parameters by approximately 54%, from 996,019 parameters in the original baseline model down to 460,247, which means a significant gain in memory efficiency.

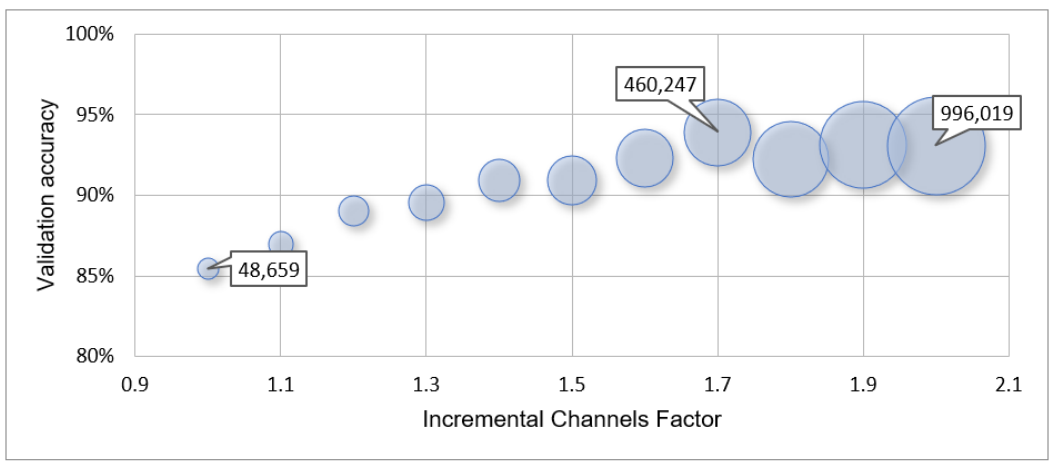

Figure 10. Change in validation accuracy as a result of reducing the number of channels between convolutional layers.

The next strategy we use to lessen the complexity of the baseline model focuses on the gradual reduction of the number of neurons in the classification part of the model. The original baseline model uses a combination of 48 and 24 neurons in the first and second fully connected layers, and 3 neurons in the last classification layer, which adds up to a total of 75 neurons. We note that the number of neurons in the last layer must be kept equal to the number of categories in the classification task, which is 3 for determining RSC (bare pavement, partial cover, full cover).

On each consecutive experiment, we eliminate 6 neurons in the first fully connected layer and 3 neurons in the second, until we end up with a combination of 6, 3, and 3 neurons for a total of 12 units in the classification part of the model. Figure 11 shows how the accuracy is affected by gradually reducing the number of neurons from a total of 75 down to 12, along with the correspondent reduction in the total number of parameters in the model represented by the area of the circles. The validation accuracy drops drastically when the total number of neurons goes below 30; therefore, we move on using a total of 39 units in order to keep the accuracy above 90%, which left us with a total of 301,727 parameters.

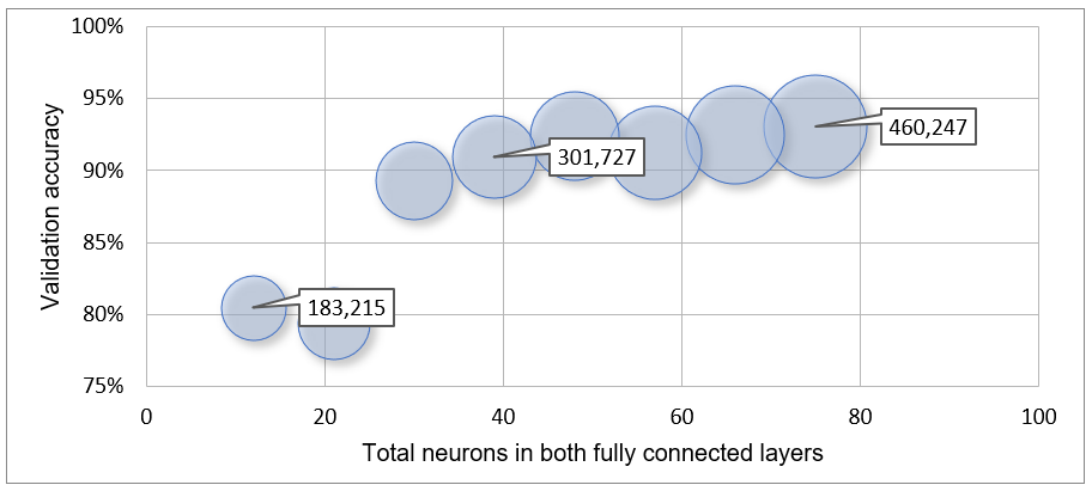

Figure 11. Change in validation accuracy as a result of reducing the number of neurons in the classification part of the model.

The results of the ablation study allow us to lower the complexity of the baseline model without a significant reduction in validation accuracy. This *simplified* version of the baseline model includes about 30% the number of parameters compared to the original one and only 1.3% when compared to the average number of parameters in the six state-of-the-art models we evaluate in the previous section.

Even though the accuracy went from ~93% to ~91% during the ablation study, our purpose here is to explore the effects of multiple design criteria rather than scoring the maximum possible accuracy. Using DL models with a fewer number of parameters is highly desirable for reducing training time during the design stage and for deployment in a production environment, especially for applications where input images come from hundreds of cameras every 10 to 15 minutes like in the case of determining RSC for roads in the province of Ontario.

**Complimentary use of Weather observations**

In this section, we evaluate the benefits of using weather data along with images from roadside cameras as input for monitoring RSC. RWIS stations can have different sensors depending on the vendor; however, most stations record at least the following five variables: Air Temperature (°C), Dew Point (°C), Relative Humidity (%), Pressure (kPa), and Wind Speed (km/h). For each image in the dataset (14,000) we obtain these five variables, for a total of 70,000 weather observations; however, since Dew Point is usually inferred from Air Temperature and Relative Humidity, we discard that variable for the subsequent analysis.

We concatenate the output of the *simplified* baseline model (SBM) with the values of the weather variables recorded in the same instant and location as the input image. The resulting vector includes seven features in total. About 1.7% of observations for Relative Humidity and 0.26% for Wind Speed appear as Null values; therefore, we fill those gaps with their corresponding average per feature. Z-score standardization per feature was also used. Three Machine Learning models for classification are considered: Random Forest (RF), Support Vector Machine (SVM), and Naïve Bayes (NB). For all three we use grid search and cross-validation to find an appropriate set of parameters and plot the normalized confusion matrices over the test set as shown in Figure 12.

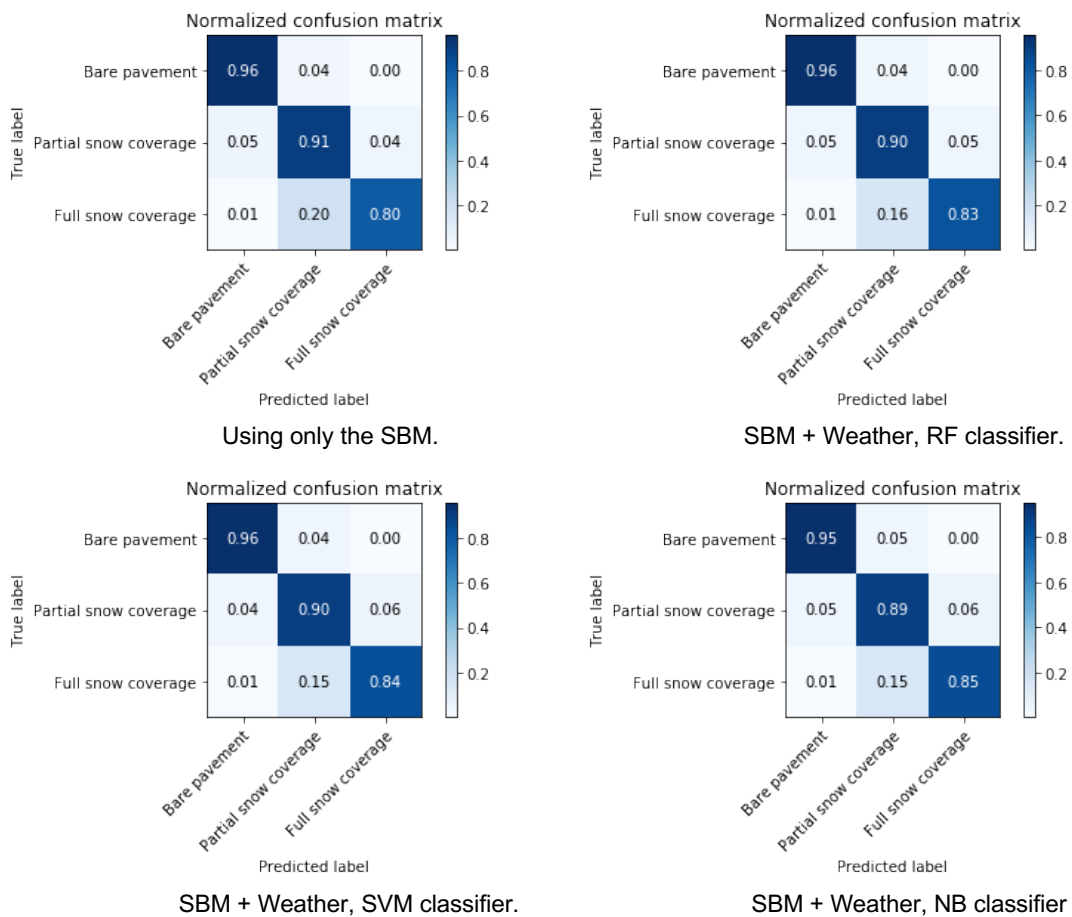

Figure 12. Normalized confusion matrices calculated over the test set.

The normalized confusion matrices show that including weather data improves the ability of the models to distinguish images with full snow coverage, with increments of 3%, 4%, and 5% in test accuracy when using the RF, SVM, and NB models accordingly. However, these improvements are in part due to a slight reduction in the classification accuracy for the other two categories. Based on the experimental results, we suggest using the Naïve Bayes classifier to achieve the highest accuracy per RSC category, although the F1-scores for the three models are very similar, with 91.78% for RF, 91.80% for SVM, and 91.24% for NB.

The optimal combination of parameters found during the grid search procedure for the three methods are as follow:
- Random Forest: Number of trees = 50, max depth = 6, and minimum samples per leaf = 4
- SVM: Kernel is radial basis function, penalty of the error term = 100, and gamma = 0.1
- Naïve Bayes: Variable smoothing = 0.01

**Conclusions and future work**

The results of this study confirm the effectiveness of DL models for determining RSC from roadside camera images. We evaluate the performance of seven DL models under multiple training and fine-tuning scenarios and select the baseline model as the one that produces the best experimental accuracy. We then explore the simplification of the baseline model through an ablation study and find that we can achieve ~91% accuracy using a model that is 98.7% smaller than recently published state-of-the-art DL models. However, we are aware that these results are only indicative of the performance of these models for our specific application. Furthermore, we find that using weather data slightly improves the classification accuracy for images in the fully covered RSC category.

While there is a direct applicability of the methods presented in this paper to improve Winter road maintenance operations, some challenges are worth to note, such as the required changes for the current technology infrastructure in order to accommodate the new software components running these automated tasks, as well as the introduction of visualization dashboards in the monitoring centres for the aggregation of results.

We highlight the value of the experimental results presented in this study toward the automated detection of RSC for Winter road monitoring, not only for improving road safety but also for resource optimization. The contributions of this paper are aligned with one of the strategic objectives of Canada's Road Safety Strategy 2025: "Leveraging technology and innovation" and two of the key road safety suggested interventions: use of technology and data-driven research [1]. Moreover, the use of Machine Learning techniques for automating data processing can complement other applications on Winter road safety such as accident prediction models [37].

Future research should focus on replicating this study in other provinces across Canada and countries in the Northern hemisphere, as well as exploring the integration of these automated methods with the software architecture of the current monitoring systems.